\documentclass{article}

    \usepackage[preprint]{neurips_2025}

\usepackage[utf8]{inputenc} 
\usepackage[T1]{fontenc}    
\usepackage{hyperref}       
\usepackage{url}            
\usepackage{booktabs}       
\usepackage{amsfonts}       
\usepackage{amsmath}        
\usepackage{nicefrac}       
\usepackage{microtype}      
\usepackage{xcolor}         
\usepackage{overpic}
\usepackage{array}
\usepackage{multirow}

\usepackage{natbib}
\setcitestyle{numbers,square}

\title{PartDexTOG: Generating Dexterous Task-Oriented Grasping via Language-driven Part Analysis}

\author{
    Weishang Wu\textsuperscript{1}, 
    \quad Yifei Shi\textsuperscript{2},
    \quad Zhizhong Chen\textsuperscript{2},
    \quad Zhiping Cai\textsuperscript{1} \\ 
    \textsuperscript{1} College of Computer Science and Technology, National University of Defense Technology, China \\
    \textsuperscript{2} College of Intelligence Science and Technology, National University of Defense Technology, China  \\
    \tt{
    wuweishang24@nudt.edu.cn \quad yifei.j.shi@gmail.com} \\
    \tt{
     chenzz@nudt.edu.cn \quad zpcai@nudt.edu.cn
    } 
}

\begin{document}

\maketitle

\begin{abstract}
Task-oriented grasping is a crucial yet challenging task in robotic manipulation.
Despite the recent progress, few existing methods address task-oriented grasping with dexterous hands.
Dexterous hands provide better precision and versatility, enabling robots to perform task-oriented grasping more effectively.
In this paper, we argue that part analysis can enhance dexterous grasping by providing detailed information about the object's functionality.
We propose PartDexTOG, a method that generates dexterous task-oriented grasps via language-driven part analysis.
Taking a 3D object and a manipulation task represented by language as input, 
the method first generates the category-level and part-level grasp descriptions w.r.t the manipulation task by LLMs.
Then, a category-part conditional diffusion model is developed to generate a dexterous grasp for each part, respectively, based on the generated descriptions.
To select the most plausible combination of grasp and corresponding part from the generated ones, we propose a measure of geometric consistency between grasp and part.
We show that our method greatly benefits from the open-world knowledge reasoning on object parts by LLMs, which naturally facilitates the learning of grasp generation on objects with different geometry and for different manipulation tasks.
Our method ranks top on the OakInk-shape dataset over all previous methods, improving the Penetration Volume, the Grasp Displace, and the P-FID over the state-of-the-art by $3.58\%$, $2.87\%$, and $41.43\%$, respectively.
Notably, it demonstrates good generality in handling novel categories and tasks.
\end{abstract}

\section{Introduction}
\label{sec:intro}
Task-Oriented Grasping (TOG), which requires selecting the optimal area of an object to grasp, such that the execution of the subsequent manipulation could be facilitated, has recently attracted extensive research attention \cite{murali2021same, kokic2020learning}.
Existing works have explored achieving TOG through building knowledge graphs \cite{murali2021same, Ardon_Pairet_Petrick_Ramamoorthy_Lohan_2019}, learning from videos with human actions \cite{palleschi2023grasp, gabellieri2020grasp}, and utilizing pre-trained foundation models \cite{tang2023graspgpt, tang2024foundationgrasp}.
The generated versatile grasping enables robots to interact with objects in human-like manners \cite{kokic2020learning, xu2021affordance} and would boost a variety of downstream applications \cite{fang2020learning, ciocarlie2014towards, dai2024interfusion}.

While substantial progress has been made in TOG with parallel grippers, less attention has been paid to TOG with dexterous hands.
Dexterous hands provide better manipulation precision and versatility, enabling robots to conduct complex manipulation tasks that parallel grippers cannot accomplish.

\begin{figure}[t]\centering
    \begin{overpic}[width=0.8\linewidth,tics=10]{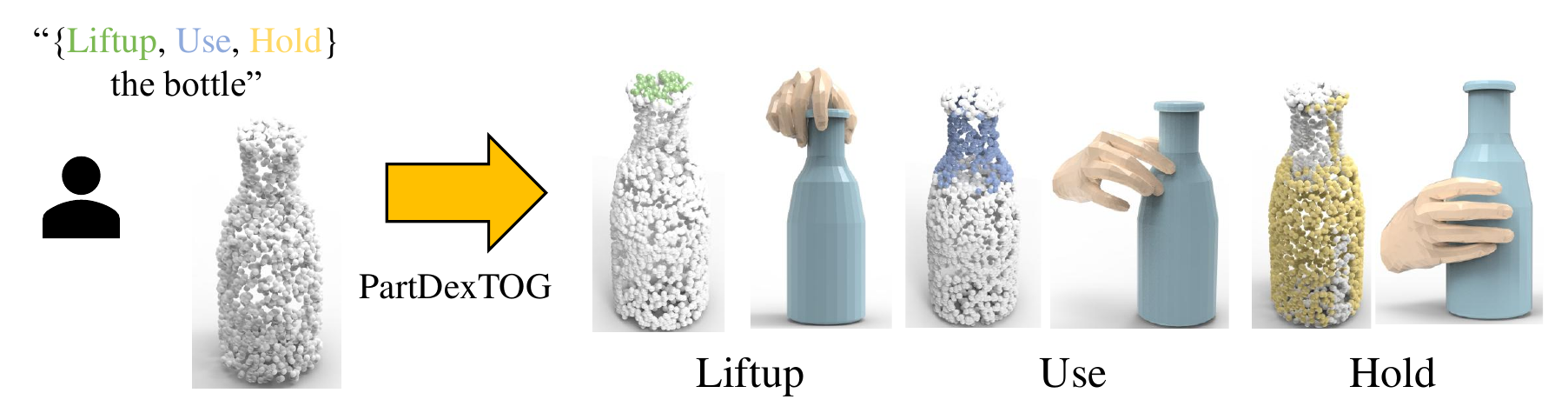}
    \end{overpic}
    \caption{PartDexTOG generates high-quality dexterous task-oriented grasps given a language instruction with the object category and a manipulation task. It selects the most plausible part of the target object to grasp, thanks to the open-world knowledge reasoning capabilities of LLMs.}
    \vspace{-8pt}
    \label{fig:teaser}
\end{figure}


Directly extending existing parallel TOG methods to dexterous hands is infeasible.
First, compared to parallel grippers, dexterous hands necessitate a higher degree of freedom, posing considerable challenges in network design, especially for estimating grasp pose on objects with novel geometry.
Second, the versatility of dexterous manipulation greatly surpasses that of manipulation by parallel grippers, rendering the datasets on task-oriented parallel grasping inadequate.
Specifically, the majority of grasp types in the Human Grasp Taxonomy \cite{feix2015grasp}, such as using scissors or chopsticks, require multiple fingers to be involved and are less likely to be conducted by parallel grippers.

In this work, we advocate generating dexterous TOG with the aid of part analysis.
Parts are fundamental elements in understanding how objects can be effectively interacted with.
On the other hand, recent works in Large Language Models (LLMs) have significantly advanced robotic manipulation, enabling zero-shot generalization in unseen scenarios \cite{xiong2024autonomous, arenas2024prompt}, thanks to the open-world knowledge reasoning capabilities of LLMs.
By incorporating LLMs, part analysis can enhance robot grasping by providing the part-level functionality of open-world objects, eliminating the necessity for training on large-scale grasping datasets.

We propose PartDexTOG, a method that generates dexterous TOG via language-driven part analysis (Figure~\ref{fig:teaser}).
Taking a 3D object and a manipulation task as input, 
our method first generates the category-level and part-level grasp descriptions w.r.t the manipulation task by LLMs.
While the category-level grasp descriptions include general grasping knowledge given the object category, the part-level grasp descriptions ground the grasping knowledge into multi-scale parts of the specific input object.
The two kinds of descriptions essentially provide information for improving the category-wise correctness and the part-wise plausibility of the generated grasps.

Then, a category-part conditional diffusion model is developed to generate a dexterous grasp for each part, based on the generated descriptions.
Specifically, the diffusion model first performs a cross-attention between the category-level grasp description and each part-level grasp description to aggregate features that contain grasp knowledge from both sides. Afterward, it generates dexterous grasps based on the aggregated features by a conditional diffusion model.
Once generating the per-part grasp, our method selects the most plausible combination of grasp and the corresponding part from the generated ones by measuring their geometric consistency, 

Our method has two main advantages. First, since it generates grasps through part analysis, it naturally facilitates grasp generation learning on objects with similar parts, greatly improving the generality to unseen objects with similar parts.
Second, as our method performs analysis on multi-scale parts, the extracted part-dependent grasp descriptions include knowledge of grasping on multi-scale regions which are beneficial to the grasp generation of objects w.r.t different manipulation tasks. This enables method to generate dexterous grasps with higher versatility compared to those with single-scale parts.

We conduct comprehensive experiments to evaluate the effectiveness of the proposed method. 
Our method ranks top on the OakInk-shape dataset over all previous methods, improving the Penetration Volume, the Grasp Displace, and the P-FID over the state-of-the-art by $3.58\%$, $2.87\%$, and $41.43\%$, respectively.
Notably, it shows good generality in handling novel categories and tasks.

In summary, the main contributions of this work are:
\begin{itemize}
\item We propose a framework that generates dexterous TOG via language-driven part analysis by leveraging LLMs' open-world knowledge reasoning capabilities.
\item We develop several LLM prompts to generate the category and part-level grasp descriptions, improving the robustness of dexterous task-oriented grasp generation.
\item We propose a diffusion model to generate dexterous TOG conditioned on grasp descriptions.
\item We achieve state-of-the-art performance on the OakInk-shape dataset, outperforming existing methods by a large margin in several evaluation metrics.
\end{itemize}

\section{Related works}
\label{sec:rw}

\subsection{Task-oriented grasping}
TOG requires detecting grasps considering the manipulation task. Early works employed analytical methods~\cite{borst2004grasp,haschke2005task, li1988task, Guan_IJCV, Guan_TCYB} to optimize task wrench space. These methods lack generality, limiting their applications. More recently, data-driven approaches have greatly boosted the area. For example, GCNGrasp \cite{murali2021same} constructs knowledge graphs to filter grasping strategies. GraspGPT \cite{tang2023graspgpt} and FoundationGrasp \cite{tang2024foundationgrasp} introduce LLMs to extract open-world semantic information. LERF-TOGO \cite{rashid2023language} generates the grasp distribution on objects utilizing LERF \cite{kerr2023lerf}. SegGrasps \cite{li2024seggrasp} utilizes vision-language models and convex decomposition to achieve object segmentation for grasping.

Several recent works adopt dexterous hands to conduct TOG, bringing better flexibility \cite{mandikal2021learning, liu2020deep}. TaskDexGrasp \cite{chen2023task} aligns task and grasp wrench space. DexGYSGrasp \cite{wei2024grasp} utilizes CLIP to achieve the analysis of task semantics. 
DexFG-Net \cite{wei2024learning} is based on a dense shape correspondence. NL2Contact~\cite{zhang2024nl2contact} generates controllable grasps by using a multi-stage diffusion model. DexTOG~\cite{zhang2025dextog} uses the intermediate generation grasp as control conditions for generating the grasp. SemGrasp \cite{li2025semgrasp} uses multi-modal LLMs to achieve cross-modal alignment and integration. SayFuncGrasp \cite{li2025language} utilizes LLMs to evaluate grasping poses and parts. AffordDexGrasp~\cite{wei2025afforddexgrasp} introduces new Generalizable-Instructive Affordance. 
TextGraspdiff \cite{chang2024text2grasp} and G-DexGrasp~\cite{jian2025g} both utilize object parts. 
Although they share a similar concept of using object parts, our method distinguishes itself by not requiring a given grasping part, which is a more general solution for downstream applications.

\subsection{LLMs for robotics}
LLMs have recently gained significant attention in the field of robotics. The unparalleled capabilities in open-world understanding and reasoning enable LLMs to boost various robotics applications \cite{shah2023lm, huang2023visual, brohan2023can}. LLMs could interpret the instructions and generate sequential actions by leveraging high-level semantic knowledge about tasks \cite{zeng2023large,wang2024large}.
These methods show good zero-shot generalization in task understanding and high-quality real-world grounding to ensure the planned actions are feasible and contextually appropriate \cite{kim2023context, huang2023instruct2act, huang2022language, huang2023inner}. Our method follows similar protocols of existing works. It explores leveraging LLMs to boost a specific task, i.e. TOG, through extracting the part functionalities from language. The proposed framework as well as the designed LLM prompts are expected to be useful for other relevant robotic tasks.

\subsection{Part analysis for grasping}
Object parts are crucial for understanding and manipulating objects. Many works leverage object parts to guide grasp generation. For example, 3DAPNet \cite{nguyen2024language} uses cosine similarity for part-specific grasp detection. LangPartGPD \cite{song2023learning} proposes a two-stage method using a language input to generate part-related grasps. ShapeGrasp \cite{li2024shapegrasp} segments objects into parts and estimates the grasp probabilities for each part. GLP2.0 \cite{palleschi2023grasp} decomposes objects to aid in novel object grasp detection. CPM \cite{liu2024composable} uses a part-based diffusion model for tool manipulation. LERF-TOGO \cite{rashid2023language} adopts LERF \cite{kerr2023lerf} to achieve zero-shot TOG with part-level LERF querying. Grasping datasets, such as AffordPose \cite{jian2023affordpose} and OakInk \cite{yang2022oakink}, were collected to provide grasp annotations as well as part segmentation of objects, enabling comprehensive evaluations of part-based grasping methods. 
Our method draws inspiration from these works but differs by detecting multi-scale object parts and learning task correspondences, enhancing flexibility and applicability across various tasks and categories.

\section{Method}
\label{sec:method} 

\subsection{Overview}
\begin{figure*}[t]\centering
    \begin{overpic}[width=0.95\linewidth]
    {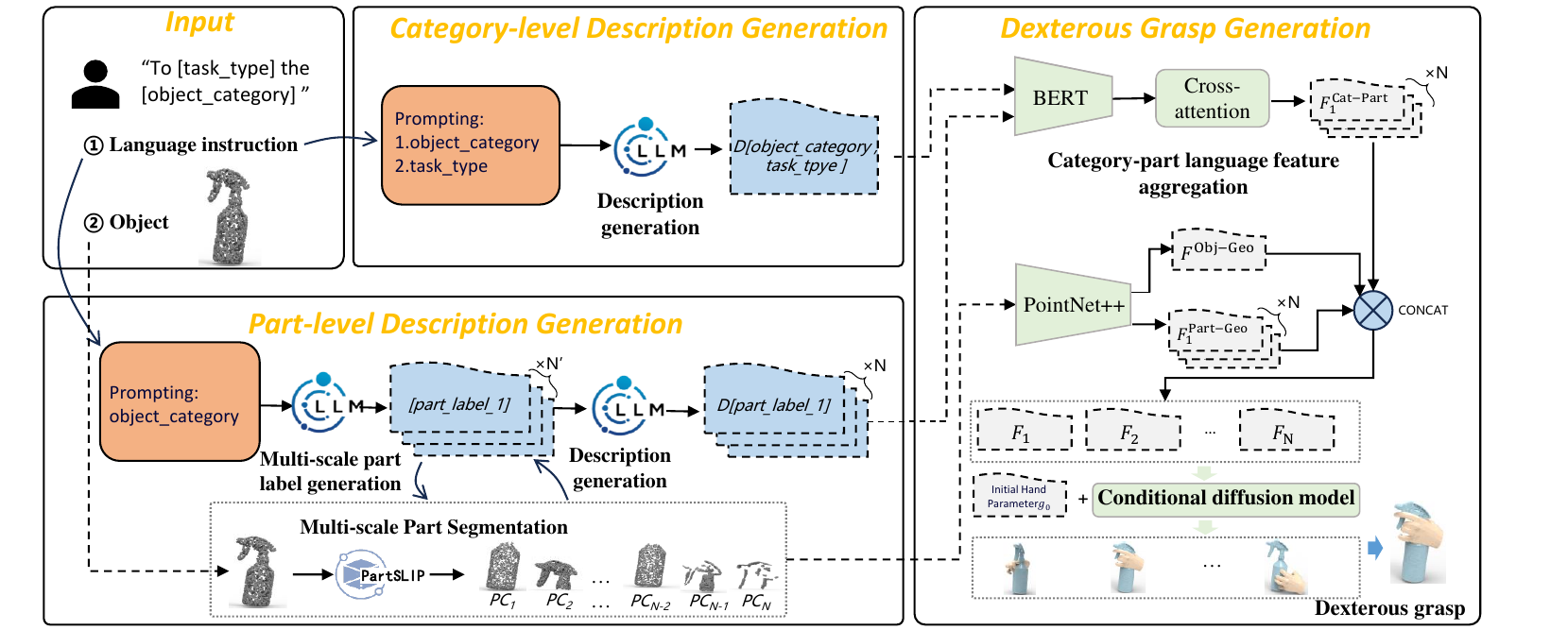}
    \end{overpic}
    \caption{\textbf{The overview of PartDexTOG}. Taking a 3D object and a manipulation task as inputs, the method first generates the category-level grasp descriptions by LLMs. Then, the potential multi-scale part labels are generated based on the object category. These part labels are segmented using PartSlip to identify the valid point sets. The part-level grasp description for each valid part is generated using LLMs.
    Last, the features of the descriptions are aggregated via cross-attention. The language features, together with the point cloud features, are fed into a conditional diffusion model to generate a dexterous grasp for each valid part, respectively. The most plausible combination of grasp and the corresponding part is selected with a measure of geometric consistency between grasp and part.}
    \label{fig:overview}
\end{figure*}

Taking a 3D object (assuming the category is known) and a manipulation task as inputs, PartDexTOG generates dexterous TOG with multiple steps.
First, it generates the category-level and part-level grasp descriptions by LLMs (Sec.~\ref{sec:description}).
Then, a category-part conditional diffusion model is developed to generate dexterous grasps for each part, based on the generated descriptions. 
The most plausible combination of grasp and corresponding part is selected by measuring the geometric consistency between the grasp and part (Sec.~\ref{sec:grasp}).
An overview is visualized in Figure~\ref{fig:overview}.

\subsection{Grasp description generation}
\label{sec:description}
\subsubsection{Category-level description generation}
\label{sec:category}
Previous algorithms typically establish correspondences between object category and task type by constructing semantic representations.
These methods only work on closed concepts and have limited generality.
To solve this problem, we opt to generate the language description for TOG by LLMs.
The open-world knowledge reasoning of LLMs brings zero-shot learning possibilities to TOG, making the solution general and effective while avoiding the need for large-scale specific TOG datasets. 

Since humans can imagine how to grasp an object to accomplish a task by only giving the object's category, the category could be deemed an important clue that includes a substantial amount of knowledge for TOG.
As such, we design a prompt considering the object category and task type. The prompt is provided in Figure~\ref{fig:prompt}(a).
In practice, we found that the generated category-level grasp description could provide general and informative knowledge that ensures the high-level grasping correctness for most typical objects in the specific category. 
We denote the category-level grasp description as $\emph{D[object\_category, task\_type]}$. 
Please refer to the appendix for the generated descriptions of representative object categories.

\subsubsection{Part-level description generation}
\label{sec:part}
Despite the above description providing the category-wise knowledge, it is still insufficient for dexterous TOG where the high degree-of-freedoms should be carefully estimated.
To address this, we introduce the part-level grasp descriptions.
Since parts are fundamental elements in understanding how objects can be interacted with effectively, 
the part-level grasp description would ground the grasping knowledge from LLMs to the detailed object geometry, which is expected to be more fine-grained compared to the category-level grasp description.
To achieve this, we first generate the category-wise multi-scale part label. Then, we determine the existence of each part label in the input object with a part segmentation algorithm. Last, the part-level descriptions are generated with LLMs. The details of the method are as follows.

\begin{figure*}[t]
\centering
    \begin{overpic}[width=\linewidth,tics=10]{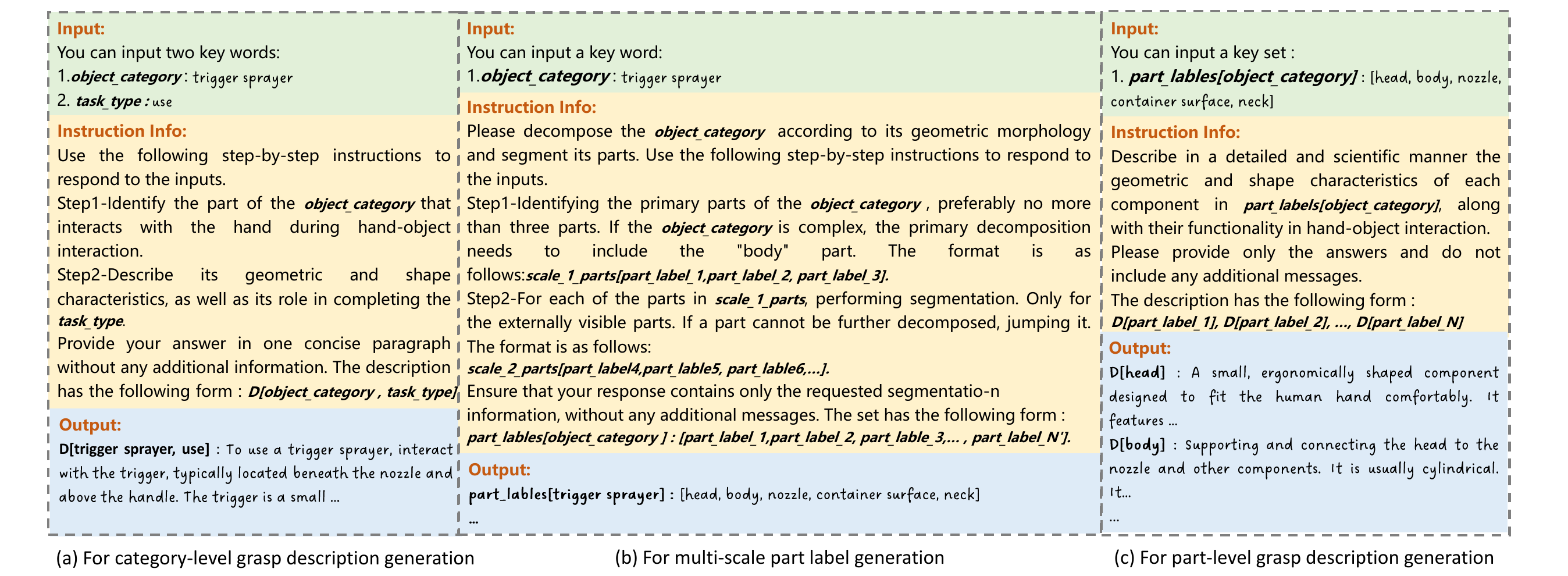}
    \end{overpic}
    \caption{The LLM prompts utilized in PartDexTOG.}
    \label{fig:prompt}
\end{figure*}

\textbf{Multi-scale part label generation.}
We first generate the potential part labels by prompting LLMs.
The number of the potential parts is denoted as $N'$.
The designed prompt is reported in Figure~\ref{fig:prompt}(b).
In particular, we consider the following factors:
1) \emph{Visibility}: only visible parts are considered;
2) \emph{Functionality}: prioritizing the parts typically involved in regions of interaction;  
3) \emph{Generality}: encouraging LLM to provide common parts, e.g., head, body, etc.
Notably, we segment object into multi-scale parts.
This makes our method compatible with more task types, significantly improving flexibility and generality.
We found that using two segmentation scales is optimal in most scenarios.

\textbf{Part segmentation.}
To ground the part labels, we adopt PartSLIP \cite{liu2023partslip} to perform part segmentation. To be specific, PartSLIP takes the point cloud and multi-scale part labels as input. It generates a set of point sets corresponding to part labels.
We then filter the point sets whose number of points is less than threshold.
The remaining point sets correspond to valid parts, $N$ is the number of valid parts. 

\textbf{Description generation.}
Once determined the valid parts, we generate part-level description with the prompt in Figure~\ref{fig:prompt}(c). To make the description more specific, the prompt includes sentences that focus on the functionalities of part considering its geometry w.r.t hand-object interaction.
It is also noteworthy that we do not take the task type as input in this prompt. 
This makes the generated part-level descriptions applicable to both the specific and general manipulation tasks, greatly improving the generality.
We denote the part-level grasp description of the $n$-th part as $\emph{D[part\_label\_n]}$.

\subsection{Dexterous grasp generation}
\label{sec:grasp}
Having generated the grasp descriptions, the next step is to compute the dexterous grasp based on them.
However, annotating the optimal part for open-vocabulary objects is non-trivial. Since the annotation of grasping part is not always available in real applications, detecting the most plausible part using supervised learning methods becomes less feasible. 
To solve this problem, we propose a method that generates the most plausible grasp and part without direct part supervision, making it more applicable.
Our solution first generates the dexterous grasp for each part based on the corresponding descriptions, respectively, and then selects the grasp by measure of geometric consistency.

\textbf{Category-part language feature aggregation.}
We first encode the category-level and part-level description by using a pre-trained BERT \cite{devlin2018bert}. We compute the BERT feature of the category-level description $F^{\text{Cat}}$ and the part-level description of the $n$-th part $F^{\text{Part}}_{n}$ as follows:
\begin{equation}
F^{\text{Cat}}=\operatorname{BERT}\left(\emph{D[object\_category , task\_type] }\right),\quad F^{\text{Part}}_{n}=\operatorname{BERT}\left(\emph{D[part\_label\_n] }\right).
\end{equation}
Note that $F^{\text{Cat}} \in \mathbb{R}^{T_{td} \times T_{d}}$, $F^{\text{Part}}_{n} \in \mathbb{R}^{T_{pd} \times T_{d}}$, where $T_{td}$ and $T_{pd}$ represent the maximum lengths of the description sequences, respectively. $T_{d}$ represents the dimension of the individual token features.

For each part, we perform a cross-attention between $F^{\text{Cat}}$ and $F^{\text{Part}}_{n}$ to aggregate the feature:
\begin{equation}
\quad F^{\text{Cat-Part}}_{n}=\operatorname{Softmax}\left(\frac{Q K_{n}^{T}}{\sqrt{d_{k}}}\right) V_{n}, n \in[1,2, \cdots, N].
\end{equation}
where $Q=\text{MLP}_Q(F^{\text{CAT}})$, $K=\text{MLP}_K(F^{\text{Part}}_{n})$, and $V=\text{MLP}_V(F^{\text{Part}}_{n})$. $\text{MLP}_Q$, $\text{MLP}_K$, and $\text{MLP}_V$ are multilayer perceptron layers. $d_{k}=T_{d}$ is the feature dimension.

We empirically found that the cross-attention would lead to better results. This is probably due to the correlation estimation mechanism which would facilitate the aggregation between the category-level and part-level features. Please refer to the experiments to see the quantitative comparison.




\textbf{Conditional diffusion model.}
We then describe the conditional diffusion model for dexterous grasp generation. To achieve this, a Denoising Diffusion Probabilistic Model (DDPM) is adopted.

The conditional network starts with feeding the point clouds of the object and $n$-th part into a PointNet++ \cite{qi2017pointnet++} respectively, resulting in the $F^{\text{Obj-Geo}}$ and $F^{\text{Part-Geo}}_n$.
Then concatenate those features:
\begin{equation}\label{eq:my_equation1}
\quad F_{n}=\operatorname{Concat}\left(F^{\text{Obj-Geo}},F^{\text{Part-Geo}}_n,F^{\text{Cat-Part}}_{n} \right).
\end{equation}
$F_{n}$ as a condition to guide the grasp generation.
We use the MANO dexterous hand \cite{romero2022embodied}.
Specifically, the hand parameters $g \in \mathbb{R}^{61}$ include the pose, shape, and palm coordinates parameters.

\begin{figure}[t]\centering
    \begin{overpic}[width=0.8\linewidth]{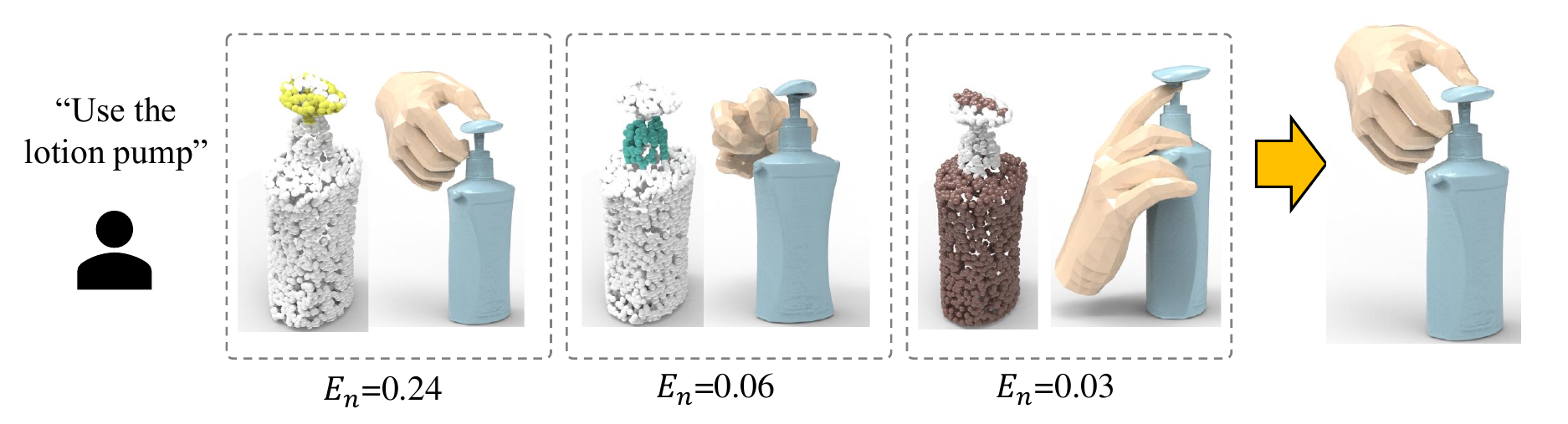}
    \end{overpic}
    \vspace{-8pt}
    \caption{Our method selects the most plausible combination of grasp and the corresponding part from the generated ones by measuring the geometric consistency between hand and part.}
    \label{fig:opt}
\end{figure}


During training, Gaussian noise is progressively added to the initial hand parameters $g_{0}$ until the distribution $g_{T}$ closely approximates a Gaussian distribution sufficiently:
\begin{equation}\label{eq:my_equation1}
q\left(g_{1: T} \mid g_{0}\right):=\prod_{t=1}^{T} q\left(g_{t} \mid g_{t-1}\right),\quad
q\left(g_{t} \mid g_{t-1}\right):=\mathcal{N}\left(g_{t} ; \sqrt{1-\beta_{t}} g_{t-1}, \beta_{t} \mathbf{I}\right),
\end{equation}
where $q \left( \cdot \right)$ is the probability distribution, $T$ is sampling step, $\beta_{1},...,\beta_{T}$ controls the noise scale. Since training process follows a Markov chain with Gaussian transitions, any step $t$ can be transformed by:
\begin{equation}\label{eq:step}
q\left(g_{t} \mid g_{0}\right):=\mathcal{N}\left(g_{t} ; \sqrt{\overline{\alpha_{t}}} g_{0},\left(1-\overline{\alpha_{t}}\right), \mathbf{I}\right), \quad\overline{\alpha_{t}}=\prod_{i=1}^{t}\left(1-\beta_{i}\right),
\end{equation}
 When $T$ is sufficiently large, an approximation to $g_{T} \sim \mathcal{~N}(0, \mathbf{I})$ can be obtained.

Loss function of DDPM is typically formulated using the Variational Lower Bound, which is commonly calculated by measuring the L1 norm between the predicted and true noise, expressed as:
\begin{equation}\label{eq:step}
 L = \mathbb{E}_{\mathbf{g}_0, \boldsymbol{\epsilon}, t} \left[ |\boldsymbol{\epsilon} - \boldsymbol{\epsilon}_\theta(\mathbf{g}_t, t, F_n)| \right],
\end{equation}
where $\mathbf{g}_0$ represents a real data sample, $\boldsymbol{\epsilon}$ is noise from the standard normal distribution, t denotes the time step, $\mathbf{g}_t$ is the noisy data at time step t, and $\boldsymbol{\epsilon}_\theta(\mathbf{g}_t, t, F_n)$ is the model's prediction of the noise.

During inference, the diffusion model $p_\theta \left( \cdot \right)$ is conditioned on the embedding features $F_{n}$ as follows:

\begin{equation}\label{eq:my_equation1}
p_{\theta}\left(g_{0: T} \mid F_n\right):=p\left(g_{T}\right) \prod_{t=1}^{T} p_{\theta}\left(g_{t-1} \mid g_{t}, F_n\right),
\end{equation}
\begin{equation}\label{eq:my_equation1}
p_{\theta}\left(g_{t-1} \mid g_{t}, F_n\right):=\mathcal{N}\left(g_{t-1} ; \epsilon_{\theta}\left(g_{t}, t, F_n\right) , \sum_{\theta}\left(g_{t}, t\right)\right), 
\end{equation}
where $g_{T}$ is the initial input which sampled from Gaussian noise, $\epsilon_{\theta}\left( \cdot \right)$ represents the denoising network implemented using a U-Net architecture, $\theta$ denotes the parameters of the diffusion model, and the variance term for Gaussian transitions can be set to $\sum_{\theta}\left(g_{t}, t\right)=\beta_{t}\mathbf{I}$.

\textbf{Measure of geometric consistency.}
Given the generated part-wise dexterous grasps, the next step is to select the most plausible grasp and the corresponding part. We found that this task could be solved by measuring the geometric consistency between grasp and part, without grasping part annotations.
To do so, we compute the area of the contact surface between the generated hand and the part:
\begin{equation}\label{eq:area_surface}
E_n=\frac{\sum_if[\min_j d(\text{PC}^\text{Part}_{n,i},\text{PC}^\text{Hand}_{n,j})]}{|\text{PC}^\text{Part}_{n}|},
\end{equation}
where $\text{PC}^\text{Hand}_n$ is the point cloud of the generated hand's surface and $\text{PC}^\text{Part}_n$ is the point cloud of the part. $d(\cdot)$ is the L2 distance. $f[\cdot]$ returns $1$ if the input value is less than a threshold $\lambda$, and returns $0$ otherwise. $|\text{PC}^\text{Part}_{n}|$ is the number of points in the part.
We select the grasp with the highest $E_n$ as the most plausible grasp. A visualization of the selection is shown in Figure~\ref{fig:opt}.


\subsection{Implementations}
We train PartDexTOG for 1000 epochs on an NVIDIA GeForce RTX 4090 GPU. The number of diffusion steps T is set as 100. 
We use the Adam optimizer with the learning rate $0.0001$ and the batch size is $64$. 
All the LLMs mentioned in our method are OpenAI's GPT-4o model~\cite{achiam2023gpt}.
Please refer to the appendix for the details of network architecture and parameter settings.

\section{Results and Evaluation}
\label{sec:result}


\begin{figure*}[t]\centering
    \begin{overpic}[width=0.95\linewidth,tics=10]{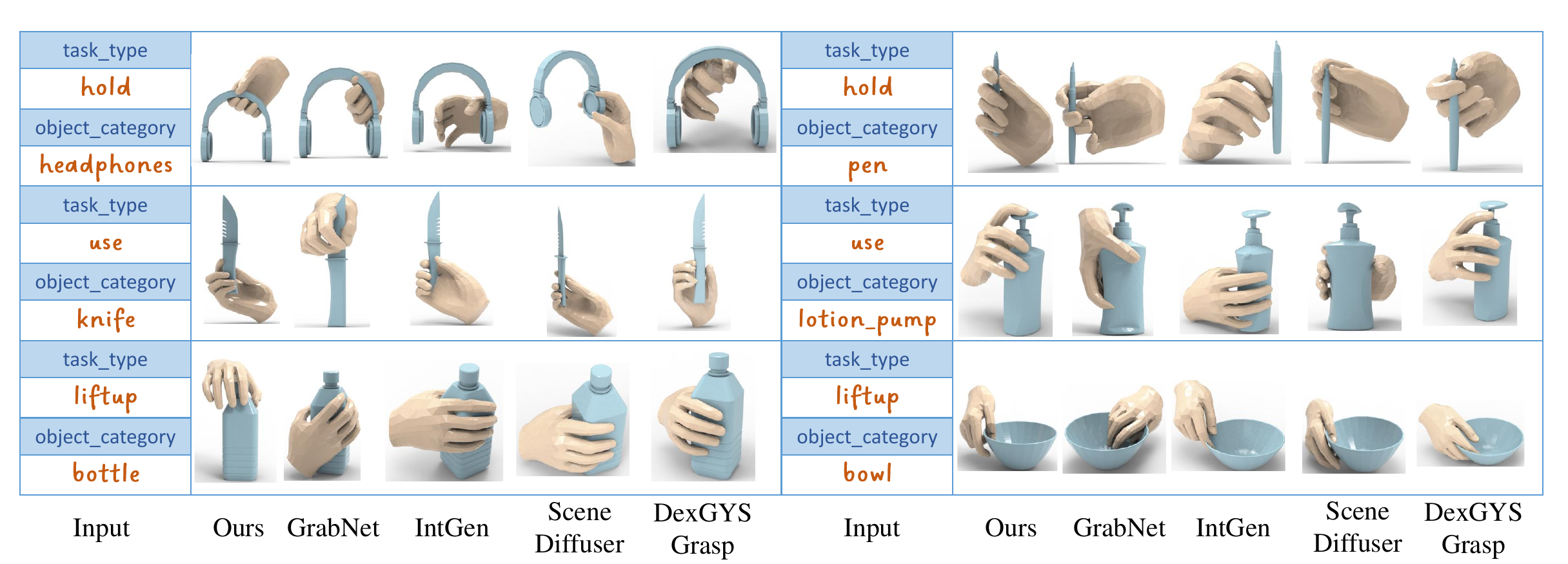}
    \end{overpic}
    \vspace{-6pt}
    \caption{Visual comparison of the generated task-oriented grasps by PartDexTOG and the baselines.}
    \label{fig:vis_baseline}
\end{figure*}
\begin{table*}[t]
    \centering
    \caption{Quantitative results on OakInk-shape.}
   
    \resizebox{1.0\linewidth}{!}{
    \begin{tabular}{c c c c c c c c c c c c c}
    \toprule
    \multicolumn{1}{c}{\multirow{2}*{Method}} & \multicolumn{2}{c}{\multirow{1}*{Penetration}} & \multicolumn{2}{c}{\multirow{1}*{Grasp Displace}} & \multicolumn{1}{c}{\multirow{2}*{Contact Radio $\uparrow$}} & \multicolumn{1}{c}{\multirow{2}*{P-FID $\downarrow$}} & \multicolumn{1}{c}{\multirow{2}*{LLM score $\uparrow$}}  & \multicolumn{1}{c}{\multirow{2}*{Diversity $\uparrow$}}& \multicolumn{4}{c}{\multirow{1}*{Perceptual Score}} \\
    {} & Volume$(cm^3) \downarrow$ & Depth$(cm) \downarrow$ &{Mean$(cm) \downarrow$} & {Var}$(cm) \downarrow$ & {}&{}&{}&{}& SC$\uparrow$& PP$\uparrow $ & IS$\uparrow $ & CP$\uparrow $  \\ 
    \midrule
    GrabNet~\cite{taheri2020grab}          & 11.61             & 0.044               & 3.91 & 4.12 & 98.84$\%$  
    & 28.28 & 44.5 & 0.671 & 3.03 & 2.75 & 2.58 & 2.38 
    \\
    IntGen~\cite{yang2022oakink}           & 10.81             & $\textbf{0.027}$ & 3.23 & 3.77& 99.32$\%$
    & 20.14 & 60.5 & 0.627 & 3.87 & 3.52 & 3.34 & 3.38 
    \\
    SceneDiffuser~\cite{huang2023diffusion}& 4.91             & 0.055 & 3.72 & 4.03 & 99.15$\%$
    & 24.66 & 62.5  & 0.984 & 3.67 & 3.87 & 3.58 & 3.27 
    \\
    DexGYSGrasp~\cite{wei2024grasp} & 10.41 
    & 0.112 & 3.34 & 3.81 & \textbf{99.73\%} & 24.14 & 64.0 & 0.715
    & 3.84  &  3.27  &  3.47  & 3.51
    \\
    \midrule
    Ours    & $\textbf{4.74}$ & 0.066               & $\textbf{3.14}$ & $\textbf{3.61}$ & $99.33\%$
    & $\textbf{14.24}$ & $\textbf{70.5}$  & $\textbf{0.993}$   & $\textbf{4.68}$ & $\textbf{4.50}$ & $\textbf{4.15}$ & $\textbf{4.41}$
    \\
    \toprule
    \end{tabular}}
    \label{tab:performance}
\end{table*}

\subsection{Dataset}
We train and test PartDexTOG on the OakInk-shape dataset \cite{yang2022oakink}, the largest dataset of dexterous TOG, containing~$50$k hand-object poses and models.
The diversities of shapes and tasks allow comprehensive evaluations of the proposed method. Specifically, the training set and testing set contains grasps on $923$ and $176$ objects, respectively. 
The latter includes objects from untrained categories, facilitating the comprehensive evaluation of generality.
This dataset is the most inclusive, diverse, and versatile one for dexterous TOG \cite{xiao2023robot}. Evaluating on this dataset only is sufficient to demonstrate the advantageous performance of our method.

\subsection{Evaluation Metrics}
To evaluate the quality of generated grasps comprehensively,
we adopt multiple evaluation metrics.
Specifically, to evaluate the physical plausibility and stability, we employ the following metrics: 
1) \textbf{Penetration}: The metrics measure the penetration.
The Penetration Volume is the volume of hand-object intersection. The Penetration Depth is the maximum penetration distance~\cite{hasson2019learning}; 2) \textbf{Grasp Displace}:
A physics simulation is applied to evaluate the grasp stability by measuring the displacement of the object's center of mass under gravity~\cite{tzionas2016capturing}; 3) \textbf{Contact Ratio}: The metric measures the hand-object interaction by sample-level contact ratio~\cite{jiang2021hand}.






Moreover, to evaluate the semantic correctness, we employ the following metrics: 1) \textbf{P-FID}: The Fréchet Inception Distance between the PointNet++ feature of the predicted and ground-truth hand~\cite{li2024seggrasp, nichol2022point}; 2) \textbf{LLM score}: A semantic score generated by the GPT-4v. We use the same setting as \cite{li2024seggrasp}; 3) \textbf{Diversity}: The variance of the grasp parameters. Higher variances represent more diversity~\cite{wang2024single}; 4) \textbf{Perceptual Scores}: We conduct user studies to evaluate the generated grasps. 
Each volunteer is asked to evaluate $500$ individual grasps, assigning perceptual scores between $0$-$5$ in the following aspects: a) Semantic Consistency (SC) between the task and the grasp; b) Physical Plausibility (PP) of hand posture; c) Interaction Stability (IS) between hand and object; d) Correctness of Part selection (CP).

 \begin{figure}[t]\centering
    \begin{overpic}[width=0.8\linewidth,tics=10]
    {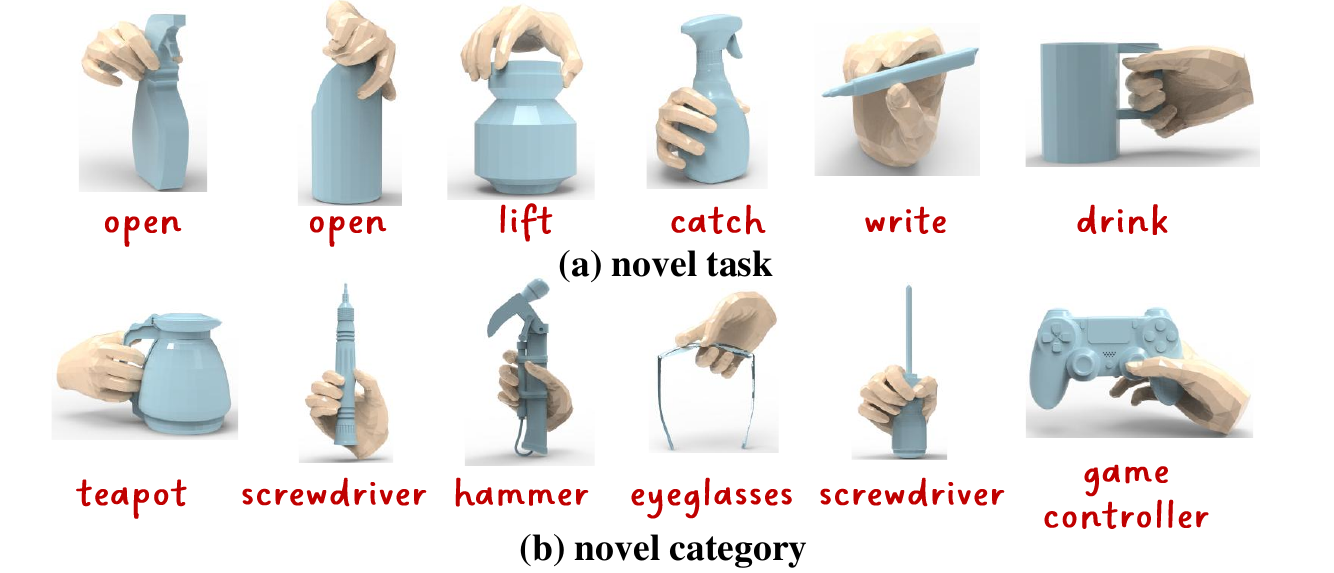}
    \end{overpic}\vspace{-8pt}
    \caption{Visualizations of the generated grasps on novel tasks (a) and categories (b). The zero-shot generalization capability of our method stems from the mechanism of language-driven part analysis.}
    \label{fig:novel}
\end{figure}

\begin{table}[t]
    \caption{Quantitative comparisons on novel tasks and category (lower is better).}
    \centering
    \resizebox{0.8\linewidth}{!}{
    \begin{tabular}{c| c c|c c}
    \toprule
    \multicolumn{1}{c}{\multirow{2}*{}} & \multicolumn{2}{c}{\multirow{1}*{Novel Tasks}} & \multicolumn{2}{c}{\multirow{1}*{Novel Category}} \\
    \midrule
    \multicolumn{1}{c|}{{Method}} & \small Penetration Volume$(cm^3) $$\downarrow$ &\small Grasp Displace$(cm)$$\downarrow$&  \small Penetration Volume$(cm^3)$$\downarrow$&\small Grasp Displace$(cm)$$\downarrow$\\
    \midrule
   GrabNet~\cite{taheri2020grab}   
     & 11.88 & 3.95 & 18.51 & 4.75 \\
    IntGen~\cite{yang2022oakink}
    & 9.71 & 3.54 & 20.04 & 3.43 \\
    SceneDiffuser~\cite{huang2023diffusion}   
    & \textbf{4.95} & 3.73 & 17.41 & 2.97 \\
    DexGYSGrasp~\cite{wei2024grasp}
    & 10.49 & 3.19 & \textbf{15.08} & 3.69 \\
    \midrule
    Ours
     & 5.30 & \textbf{3.09} & 16.22 & \textbf{2.73}\\
    \toprule
    \end{tabular}}
    \label{tab:novel}
\end{table}

\subsection{Compare to baselines}
We compare our method with the state-of-the-art on OakInk-shape. 
A detailed discussion of the baseline selection is provided in the appendix.
The quantitative results are provided in Table~\ref{tab:performance}. 
Our method outperforms all baselines in the Grasp Displace and Contact Radio metrics and achieves comparable results in the Penetration metrics, suggesting the high physical plausibility of our method.
Note that IntGen~\cite{yang2022oakink} achieves better performance in the Penetration Depth due to its conservative strategy to avoid hand-object penetration. 
Our method, in contrast, balances hand-object penetration and grasp stability better. 
For semantic consistency, Perceptual Scores highlight our method's advantages in semantic consistency, physical plausibility, interaction stability, and part selection accuracy. 
Our method also achieves the best performance on the P-FID and LLM-assisted metrics, indicating effective task compliance supported by grasp description analysis and measurement of geometric consistency.
The Diversity metric also demonstrates that our method excels in diversity.
The qualitative comparisons are in Figure~\ref{fig:vis_baseline}.

\subsection{Zero-shot generality}
We evaluate the zero-shot generality on untrained tasks and categories. Specifically, we select $5$ novel tasks and $47$ objects from $5$ novel categories for comparisons. As shown in Table~\ref{tab:novel}, our method outperforms all the baselines, indicating that our method greatly benefits from the language-driven part analysis that brings open-world knowledge reasoning for grasping generation. The visualizations are provided in Figure~\ref{fig:novel}.
More examples can be found in the appendix.

\subsection{Effects of part selection}
\begin{figure}[t]\centering
    \begin{overpic}[width=0.8\linewidth,tics=10]{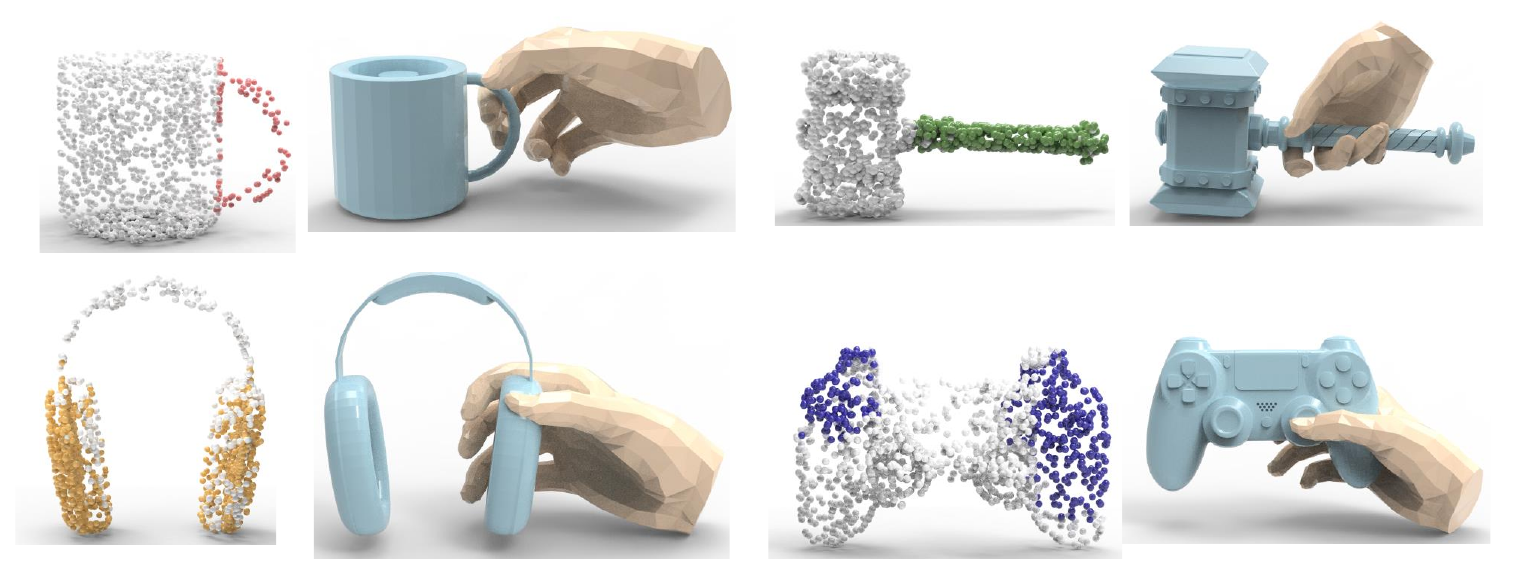}
    \end{overpic}\vspace{-8pt}
    \caption{Effects of part selection. Our method could adopt different parts to generate plausible grasps for different objects with the same input task type (use), even there is a failed part segmentation.}
    \label{fig:part_task_corr}
\end{figure}

In Figure~\ref{fig:part_task_corr}, we visualize the selected parts and generated grasps for objects in different categories where the task type is all ``use''.
It shows that our method could adopt different parts to generate plausible grasps, implying the strong capabilities in part selection. At the same time, the method is able to generate correct grasps despite the failure of the part segmentation algorithm.  
The detailed statistics of the task-part correspondence produced by our method and the impact of the segmentation results on the performance of the method can be found in the appendix.

\subsection{Ablation studies}

In Table~\ref{tab:abs}, we conduct ablation studies to quantify the efficacy of several crucial components.


\begin{table*}[t]
    \caption{Ablation studies of several crucial components in PartDexTOG.}

    \centering
    \resizebox{1.0\linewidth}{!}{
    \begin{tabular}{c c c c c c c c c c c c c}
    \toprule
    \multicolumn{1}{c}{\multirow{2}*{Method}} & \multicolumn{2}{c}{\multirow{1}*{Penetration}} & \multicolumn{2}{c}{\multirow{1}*{Grasp Displace}} & \multicolumn{1}{c}{\multirow{2}*{Contact Radio $\uparrow$}} & \multicolumn{1}{c}{\multirow{2}*{P-FID $\downarrow$}} & \multicolumn{1}{c}{\multirow{2}*{LLM score $\uparrow$}}  & \multicolumn{1}{c}{\multirow{2}*{Diversity $\uparrow$}}& \multicolumn{4}{c}{\multirow{1}*{Perceptual Score}} \\
    {} & Volume$(cm^3) \downarrow$ & Depth$(cm) \downarrow$ &{Mean$(cm) \downarrow$} & {Var}$(cm) \downarrow$ & {}&{}&{}&{}& SC$\uparrow$& PP$\uparrow $ & IS$\uparrow $ & CP$\uparrow $  \\ 
    \midrule

    \emph{ w/o category-level grasp description}
    & 11.16 & 0.074 & 4.72 & 4.53  & 88.24$\%$
    & 22.95 & 36.5  & 0.874 & 2.93 & 2.35 & 2.54 & 2.84\\
    \emph{w/o part-level grasp description}
    & 6.05 & 0.071 & 3.64 & 3.97  & 96.68$\%$
     & 14.98 & 51.5  & \textbf{1.025}& 3.41 & 3.37 & 3.09 & 3.53
    \\
    \emph{w simpler grasp description}
    & 5.02 & 0.087 & 3.29 & 3.70  & \textbf{99.42}$\%$
     & 14.71 & 68.5  & 0.948& 4.45 & 4.29 & 4.09 & 4.23\\
   
    \emph{w/o multi-scale parts}
    & \textbf{4.57}  & 0.067 & 3.40 & 3.69 & 97.59$\%$
    & 15.67 & 57.5  & 0.985& 4.12 & 3.67 & 3.78 & 4.02 \\
    \emph{w/o cross-attention}
    & 5.87  & 0.076  & 3.47 & 3.80  & 99.06$\%$
     & 14.81 & 58.5  & 0.960& 3.63 & 3.57 & 3.22 & 3.98\\
    \emph{w/o measure of geometric consistency}
    & 6.61  & 0.083  & 4.72 & 4.77  & 98.33$\%$
     &16.72 & 39.5  & 0.974& 3.12 & 2.84 & 2.61 & 2.76 \\
    \midrule
    The full method
    & 4.74 & \textbf{0.066} & \textbf{3.14} & \textbf{3.61}& 99.33$\%$
    & \textbf{14.24} & \textbf{70.5}  & 0.993& $\textbf{4.68}$ & $\textbf{4.50}$ & $\textbf{4.15}$ & $\textbf{4.41}$\\
    \toprule
    \end{tabular}}
    \label{tab:abs}
\end{table*}

\textbf{Category-level grasp description.} 
The category-level description contains general knowledge of typical object grasping.
We remove its feature (\emph{w/o category-level grasp description}), resulting in a clear performance drop.
The result shows that the category-level description is indeed helpful.

\textbf{Part-level grasp description.} 
The part-level description generation is crucial. To evaluate this, we compare the full method to \emph{w/o part-level grasp description}, which removes the part-level description features. The decline in performance verifies the significance of the component.

\textbf{Alternative grasp description.}
Our method generates detailed grasp descriptions with the proposed prompts. To verify the necessity, we replace the original grasp description with simple language (\emph{w simpler grasp description}). The performance degradation with the simple setting indicates the necessity of the detailed grasp description.

\textbf{Multi-scale part segmentation.}
Our method employs multi-scale part segmentation to accommodate the shape diversity. To validate its effectiveness, we compare the full method to the baseline with only a single-level part being adopted: \emph{w/o multi-scale parts}. Unsurprisingly, the results demonstrate that the multi-scale part segmentation would greatly improve the performance.

\textbf{Language feature aggregation.} 
Our method uses cross-attention for language feature aggregation. To evaluate its necessity, a straightforward baseline is to aggregate features using simple network layers. The baseline of \emph{w/o cross-attention} replaces the cross-attention with a concatenation layer. The performance is inferior to the full method, suggesting that the cross-attention is more effective.

\textbf{Measure of geometric consistency.} 
The measure of geometric consistency module evaluates the generated part-wise grasps. To understand its impact, we employ a simple baseline by computing the cosine similarity of category-level to part-level descriptions: \emph{w/o measure of geometric consistency}. The obvious performance decrease shows the significance of this component.

\section{Conclusion}
We have presented PartDexTOG, which generates dexterous TOG via language-driven part analysis by leveraging LLMs’ open-world knowledge reasoning capabilities. Several LLM prompts are proposed to generate the category-level and part-level grasp descriptions. Based on these descriptions, a conditional diffusion model is developed to generate dexterous grasps. We have achieved state-of-the-art performance on a large-scale public dataset, outperforming the existing methods by a large margin. An interesting future direction is to explore leveraging the pre-trained vision models to further boost dexterous TOG generation on objects with open-vocabulary categories.

\bibliographystyle{IEEEtran}
\bibliography{refs.bib}

\newpage
\appendix

\section{Appendix / supplemental material}

\subsection{More Implementation Details}
We provide the details of network architecture and implementations for clarification. 

\subsubsection{Grasp description generation}
The large language models might generate different descriptions for the same prompt. To improve the robustness of the generated descriptions, during both training and testing, we generate $10$ descriptions with the sample prompt and select a random one from those. We found this trick to be crucial to our method, especially when it comes to novel tasks and objects.

For the multi-scale part segmentation, we employ PartSlip \cite{liu2023partslip} with the pre-trained GLIP \cite{li2022grounded} model to achieve zero-shot segmentation. The hyper-parameter settings of PartSlip are provided in Table~\ref{tab:partslip}.



\subsubsection{Dexterous grasp generation}
For the category-part language feature aggregation, a pre-trained BERT model \cite{devlin2018bert} for computing the feature of the category-level description $F^{Cat}\in\mathbb{R}^{T_{td}\times T_d}$ and that of the part-level description of the n-th part $F^{Part}_n\in\mathbb{R}^{T_{pd}\times T_d}$, where $T_{td}$ and $T_{pd}$ represent the maximum lengths of the description sequences, respectively. $T_d$ represents the dimension of the individual token features output by the BERT model. We perform cross-attention to aggregate the features. The associated hyper-parameters are detailed in Table~\ref{tab:cate}.


\begin{table}[htbp]
  \centering
  \begin{minipage}[b]{0.45\textwidth}
    \caption{Hyper-parameter setting of PartSlip.}
    \centering
    \begin{tabular}{c c}
        \toprule
        Hyper-parameter & Value\\ 
        \midrule
        \# Views & 12\\
        Image Width & 800 \\
        Image Hight & 800\\
        Point size & 0.005\\
        \bottomrule
    \end{tabular}
    \label{tab:partslip}
    
  \end{minipage}
  \hfill
  \begin{minipage}[b]{0.45\textwidth}
    \caption{Hyper-parameter setting of category-part language feature aggregation network.}
    \centering
    \begin{tabular}{c c}
        \toprule
        Hyper-parameter & Value\\ 
        \midrule
        $T_{td}$ & 200\\
        $T_{pd}$ & 200\\
        $T_{d}$ & 768\\
        Dimension in cross-attention & 128\\
        \bottomrule
    \end{tabular}
    \label{tab:cate}
  \end{minipage}
  \hfill
\end{table}


    


\begin{figure}[htbp]
\centering
    \begin{overpic}[width=0.8\linewidth]{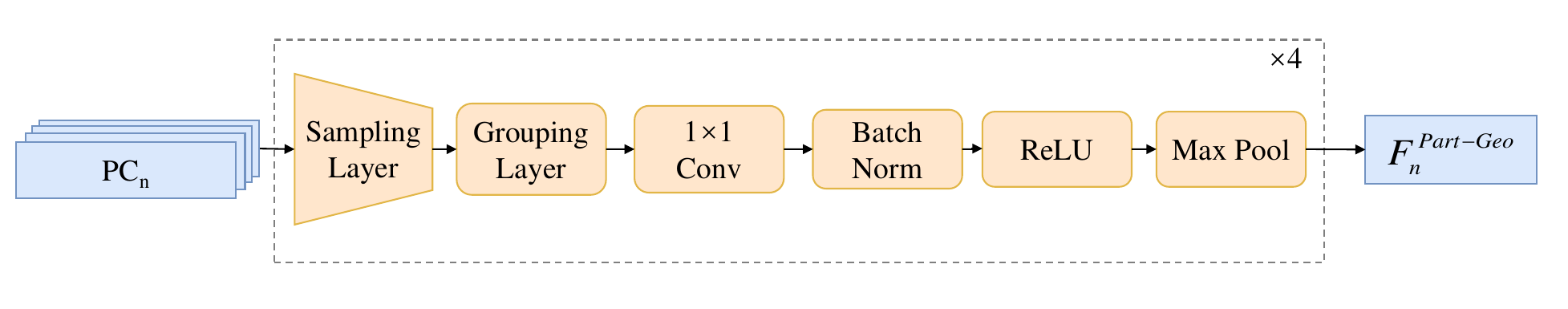}
    \end{overpic}
    \vspace{-10pt}
    \caption{The architecture of the Pointnet++.}
    \label{fig:pointnet}
\end{figure}

We use Pointnet++~\cite{qi2017pointnet++} to extract features from point clouds of the object and the parts. The network architecture is shown in Figure~\ref{fig:pointnet}, and the specific hyperparameter settings are provided in Table~\ref{tab:pointnet}.

For the conditional diffusion model, the noise schedule scales $\beta_1,\beta_2,...,\beta_T$ linearly from 0.0001 to 0.01. The model is trained for 1000 epochs with a batch size of 64, using the Adam optimizer with a learning rate of $1\times 10^{-4}$. The hyper-parameters are detailed in Table~\ref{tab:diffusion}.



\begin{figure*}[t]
\centering
    \begin{overpic}[width=\linewidth]{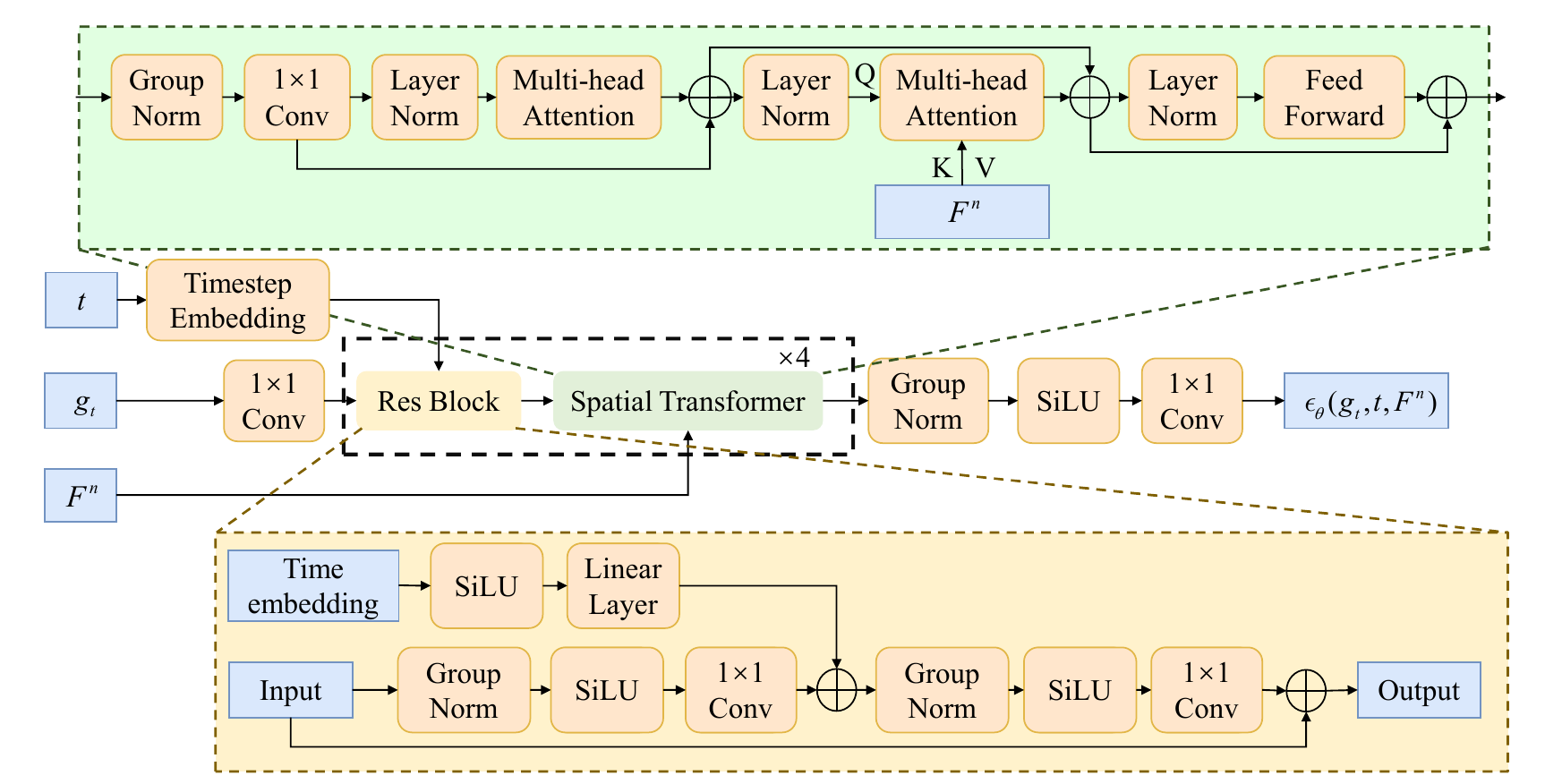}
    \end{overpic}
    \caption{The architecture of the U-Net.}
    \label{fig:unet}
\end{figure*}

The conditional diffusion model's noise prediction network is implemented using a U-Net architecture \cite{ronneberger2015u}. Initially, the time step $(t_s)$ is converted to an embedding vector via the time embedding module. Subsequently, input features are transformed into high-dimensional representations by the initial convolutional layer. These representations are refined through four sequential processing modules, each comprising a ResBlock and a Spatial Transformer. The ResBlock enhances feature representation through residual connections and feature integration, while the Spatial Transformer captures long-range dependencies through self-attention and provides conditional control via the conditional feature. After processing through all modules, the features are generated by a convolutional layer, producing the denoised target data. The U-Net architecture can be found in Figure~\ref{fig:unet}.

\begin{table}[htbp]
  \centering
  \begin{minipage}[b]{0.48\textwidth}
    \caption{Hyper-parameter setting of the PointNet++. }
    \centering
    \begin{tabular}{c c}
        \toprule
        Hyper-parameter & Value\\ 
        \midrule
       \# of SA Layer & 4\\
       \# of Sampled Points& 1024, 256, 64, 16\\
        Embedding Sizes & 64, 128, 256, 512\\ 
        \bottomrule
    \end{tabular}
    \label{tab:pointnet}
    
  \end{minipage}
  \hfill
  \begin{minipage}[b]{0.48\textwidth}
    \caption{Statistics of task-part correspondence.}
   \centering
   \begin{tabular}{c c c c}
       \toprule
        & Body & Head & Neck \\ 
       \midrule
       Hold & ${60.65\%}$ & ${29.22\%}$ & ${10.11\%}$ \\ 
       Liftup  & ${17.35\%}$ & ${67.71\%}$ & $14.92\%$ \\
       Use & $15.92\%$ & ${52.60\%}$ & ${31.48\%}$ \\
       Open & 20.92\% & ${48.34\%}$ & ${30.73\%}$ \\
       \bottomrule
   \end{tabular}
   \label{tab:part_pre}
  \end{minipage}
  \hfill
\end{table}

\subsubsection{User Study}
In the experiment, in order to compare the performance of the crawls generated by various algorithms, we invited 7 volunteers to rate the crawls generated by each algorithm. Each algorithm will uniformly extract a total of 500 identical objects and intentions. And provide the following textual explanations:

Scoring Instructions

Please ask the volunteers to rate the grasping quality, mainly evaluating the following four aspects. The higher the score, the better:

1) The provided tasks are semantically consistent with the captured ones. (0-5 points, the higher the better)

2) Physical rationality of hand posture (0-5 points, the higher the better)

3) Stability of interaction between hands and objects (0-5 points, the higher the better)

4) Whether the captured Part is reasonable (0-5 points, the higher the better)

When conducting the statistics, please provide a separate score for each item respectively.

\subsection{Additional Experiments}
\subsubsection{Statistics of the task-part correspondence}
In Table~\ref{tab:part_pre}, we provide detailed statistics on the percentages of the task-part correspondence produced by our method over three categories, i.e. bottles, lotion pumps, and trigger sprayers. Note that while the Body and Head belong to the parts of the first scale, the Neck corresponds to the parts of the second scale. The results show that our method could automatically select the proper parts to grasp, demonstrating the flexibility of our method.

\begin{figure}[ht]
\centering
    \begin{overpic}[width=\linewidth]{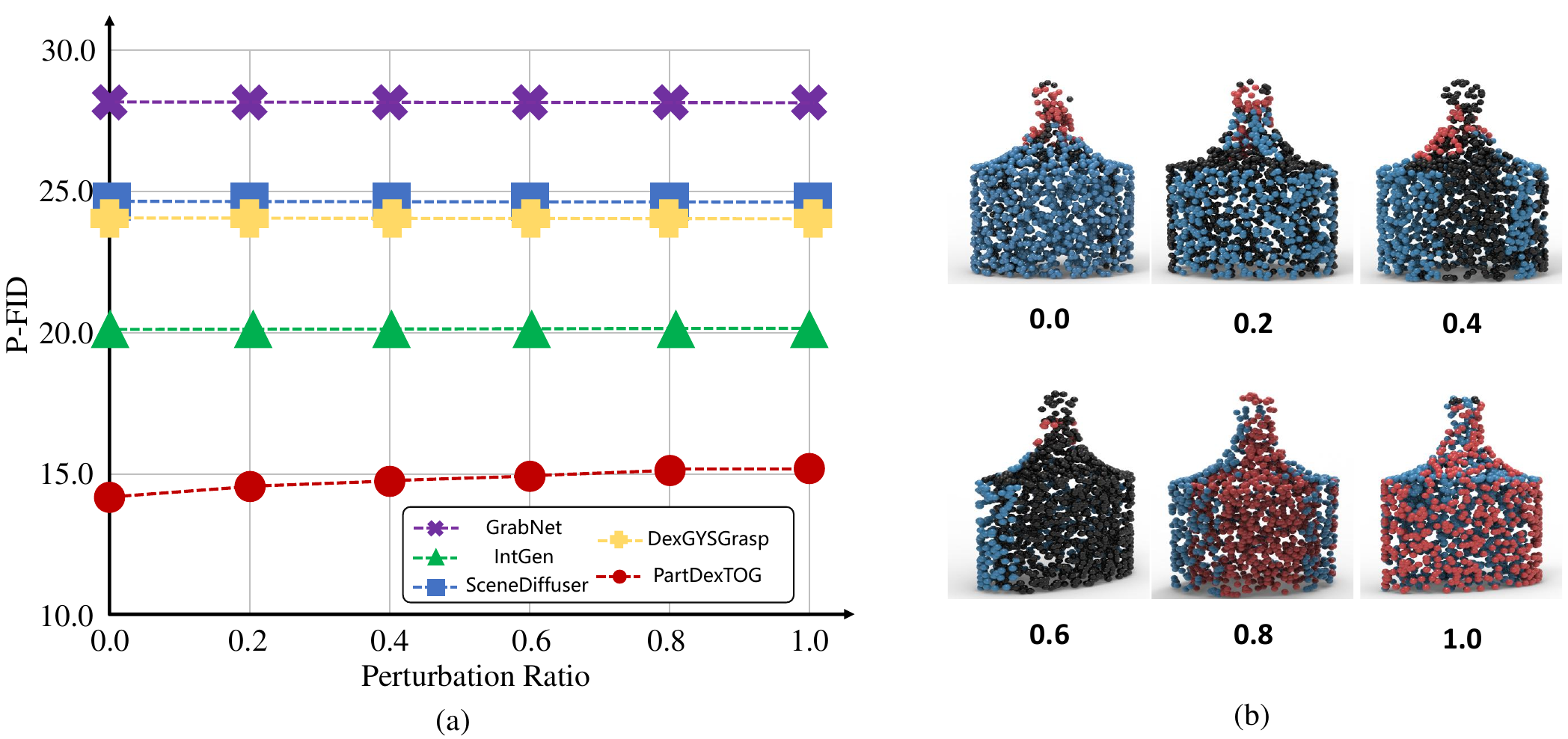}
    \end{overpic}\vspace{-8pt}
    \caption{(a) The P-FID at different levels of perturbation. (b) The examples of segmentation at different levels of perturbation.}
    \label{fig:balance}
\end{figure}

\begin{table}[t!]
    \caption{Hyper-parameter setting of the conditional diffusion model.}
    \centering
    \begin{tabular}{c c}
        \toprule
        Hyper-parameter & Value\\ 
        \midrule
       Diffusion Steps $T$ & 100\\
       Diffusion noise schedule $\beta_t$ & Linear\\
        Start value $\beta_0$ & 0.0001\\
        End value $\beta_T$ & 0.01\\
        Time embedding & Sinusoidal\\
        Transformer number of head & 8\\
        Transformer hidden dim & 64\\
        Transformer dropout & 0.1\\       
        Feed-forward network hidden dim & 128\\
        \bottomrule
    \end{tabular}
    \label{tab:diffusion}
\end{table}

\subsubsection{Performance under imperfect segmentation}
In our method, we employ the PartSlip \cite{liu2023partslip} as the method for part segmentation. However, PartSlip does not always produce flawless segmentations for objects, especially when it comes to novel shapes. To further investigate the performance of our method under imperfect part segmentation, we manually add perturbations to the output of the PartSlip. Specifically, we adopt two strategies. First, we randomly assign an incorrect part label to the patches that lie near the segmentation boundaries, since these patches are more error-prone.
The patches are generated in the PartSlip algorithm.
Second, We replace the original GLIP-L in PartSlip with the more compact GLIP-T ones, resulting in inferior part segmentations.
We generate a large number of imperfect segmentation and categorize them into five subsets with different perturbation levels.  An example can be found in Figure~\ref{fig:balance}(b).
Then, we use our method to generate the grasp descriptions and task-oriented grasps by using the imperfect segmentations.
The results are reported in Figure~\ref{fig:balance}(a). It shows that our method is generally stable with different levels of perturbation.
Moreover, our method outperforms all the baselines even with relatively high perturbations.
These show the robustness of our method.

\begin{figure}[htbp]
  \centering
  \begin{minipage}[b]{0.45\textwidth}
    \centering
    \begin{overpic}[width=\linewidth]{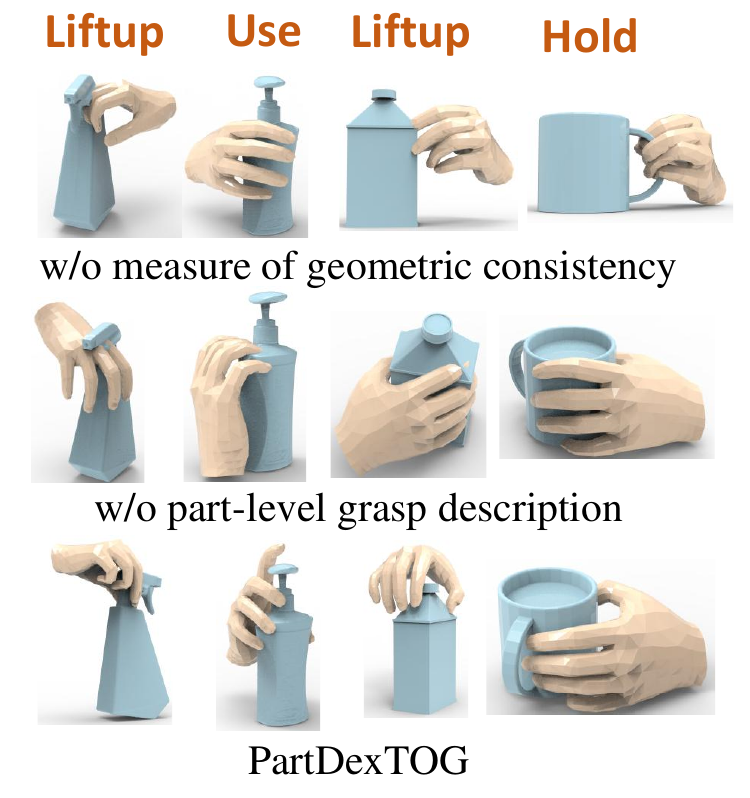}
    \end{overpic}
    \caption{Visual comparison of the generated task-oriented grasps.}
    \label{fig:abs}
  \end{minipage}
  \hfill
  \begin{minipage}[b]{0.45\textwidth}
    \centering
    \begin{overpic}[width=\linewidth]{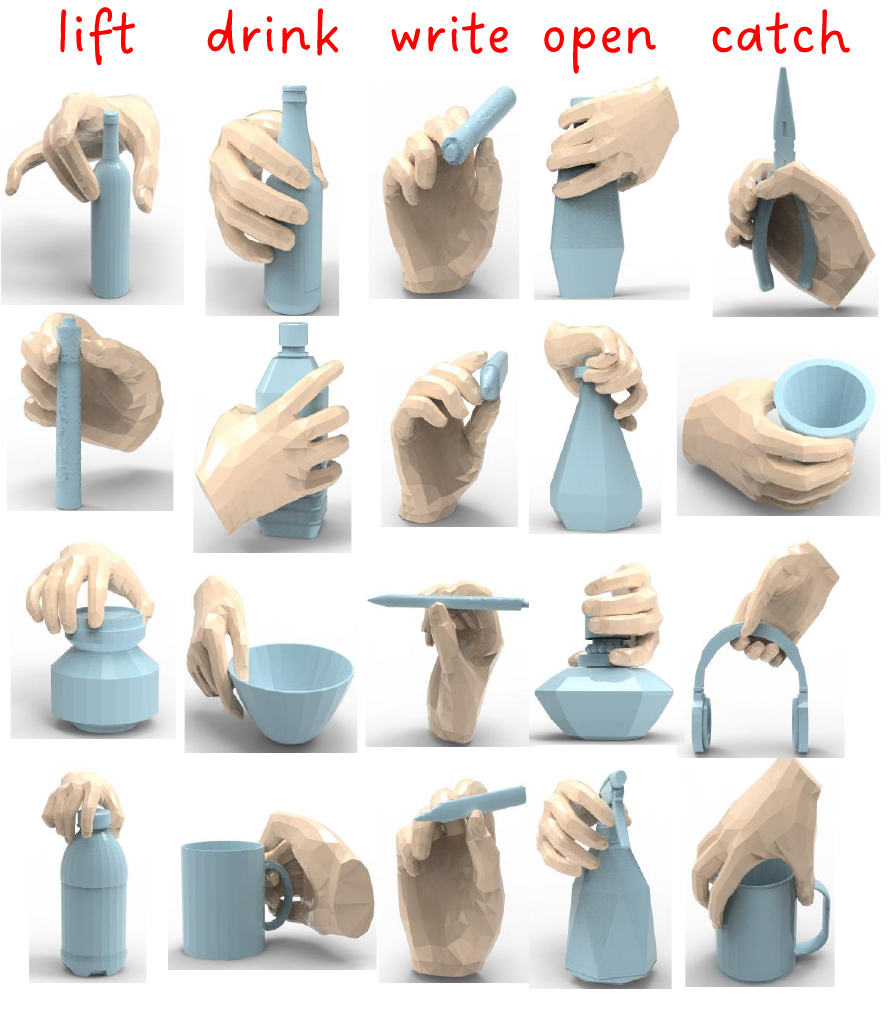}
    \end{overpic}
    \caption{Additional visualizations of the generated grasps on novel tasks.}
    \label{fig:task}

  \end{minipage}
\end{figure}

\subsubsection{Additional visualizations}
\begin{figure*}[h]
\centering
    \begin{overpic}[width=0.93\linewidth]{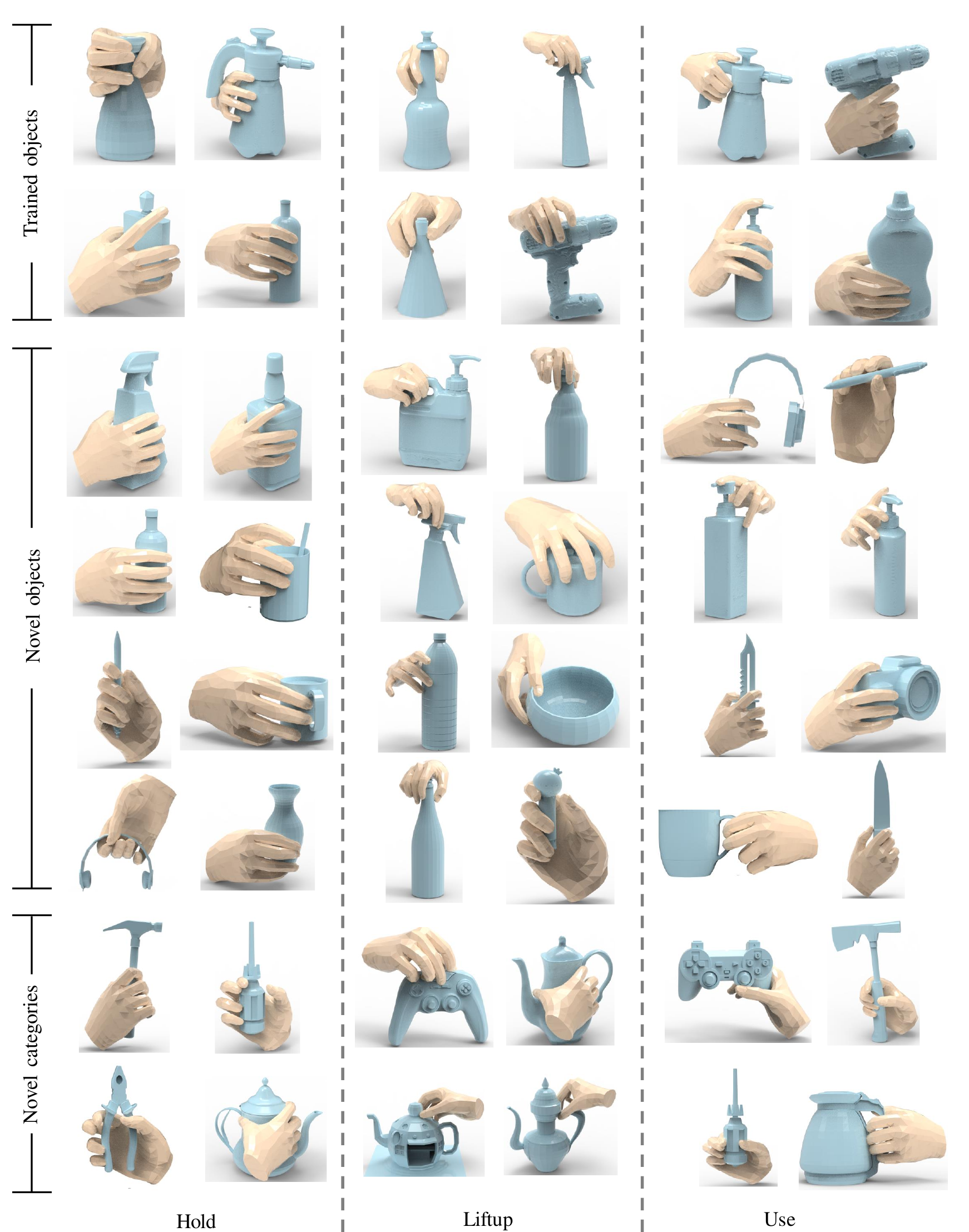}
    \end{overpic}
    \caption{Additional visualization of generated dexterous grasps on trained instances, novel instances, and novel categories.}
    \label{fig:all}
\end{figure*}

More visualizations of the PartDexTOG results are shown in Figure~\ref{fig:all}, where we showcase the results for trained instances, novel instances, and novel categories separately. More visualizations of the grasping on novel tasks are shown in Figure~\ref{fig:task}. More visualisation comparison in partial ablation experiments can be found in Figure~\ref{fig:abs}

\subsection{Discussion of Baseline Methods in Task-oriented Dexterous Grasping}
Our method generates dexterous task-oriented grasping from a given task type and a 3D object. As discussed in the introduction and related work sections, few existing works address this challenging task. Several concurrent works study a similar problem. We did not compare with those methods due to the significant difference in problem settings~\cite{feng2024dexgangrasp,zhao2024dexh2r,chen2023task, wei2025afforddexgrasp} or the unreleased source code/training data~\cite{zhao2024dexh2r, jian2025g, zhang2024nl2contact}.



\subsection{Examples of the Generated LLM Descriptions}
We provide several representative examples of the generated category-level descriptions, part-level descriptions, and multi-scale part labels in 
Table~\ref{tab:part} - Table~\ref{tab:part_lotion}.

\subsection{Societal Impacts and Limitations}
The core innovation of this paper has a positive impact on society. We propose a PartDexTOG, a novel algorithm for generating dexterous task-oriented grasping through language-driven part analysis. It can serve as a development in fields such as human-computer interaction and provide a positive impact for robots to perform tasks in complex scenarios.

However, our method still faces some challenges. For example, for the part segmentation algorithm, it is feasible to choose a better algorithm. Meanwhile, our algorithm is currently only capable of generating a single dexterous hand grab. In future work, we will further improve the quality of the TOG grasp generation.

\begin{table*}[b]
    \caption{Representative results of the generated multi-scale part label.}
    \centering
     \renewcommand{\arraystretch}{1.5}
    \begin{tabular}{c p{0.8\textwidth}}
    \toprule
    Object\_category & \multicolumn{1}{c}{Part\_lables[object\_category]} \\
    \midrule
    Bottle & [``body", ``lid", ``body surface", ``neck", ``rim"] \\
    Bowl & [``body",``rim", ``surface", ``bottom"] \\
    Cameras & [``body", ``lens", ``buttons", ``surface", ``lens\_body", ``glass", ``display"]\\
    Cup&	[``body", ``rim", ``surface", ``base"]\\
Cylinder\_bottle	&[``body", ``head", ``surface", ``neck", ``cap"]\\
Headphones	& [``body", ``cap", ``band", ``cable", ``mic"]\\
Knife	&[``blade", ``handle", ``edge", ``spine", ``tip", ``grip", ``butt"]\\
Lotion\_pump	 &[``head", ``body", ``nozzle", ``neck", ``container surface"]\\
Mug	&[``body", ``handle", ``rim", ``side surface", ``base"]\\
Pen	&[``body", ``head", ``clip", ``barrel", ``grip", ``nib", ``sleeve", ``end plug"]\\
Pincer	&[``handle", ``pivot", ``jaws", ``hinge", ``tip", ``blade"]\\
Power\_drill	&[``body", ``handle", ``trigger", ``chuck", ``switch", ``chuck key"]\\
Scissors	&[``blade", ``handle", ``Blade edge", ``Blade surface", ``Finger loop", ``Handle grip", ``Pivot screw"]\\
Squeezable	&[``body", ``head", ``body surface", ``neck", ``spout", ``grip\_section"]\\
Trigger\_sprayer&	[``head", ``body", ``nozzle", ``container surface", ``neck"]\\
Wrench&	[``handle", ``jaw", ``slider", ``movable jaw", ``fixed jaw", ``adjustment screw"]\\

    \bottomrule
    \end{tabular}
    \label{tab:part}
\end{table*}

\begin{table*}[b]
    \caption{The generated category-level description of Bottle.}
    \centering
     \renewcommand{\arraystretch}{1.5}
    \begin{tabular}{c p{0.8\textwidth}}
    \toprule
    Task type & \multicolumn{1}{c}{Description} \\
    \midrule
    Hold & To complete the task of holding a bottle, the part of the object to interact with is the cylindrical body of the bottle. This section typically has a consistent and ergonomic shape that conforms to the natural grip of a human hand, often with a smooth or slightly textured surface to prevent slipping. The role of the cylindrical body in this task is to provide a stable and secure area for the hand to grasp, ensuring the bottle can be held firmly and comfortably without risk of dropping. \\
    Liftup & To complete the task ``liftup" with a bottle, interact with the grip area, typically the middle section or body of the bottle, which is cylindrical and smooth in shape. This part of the bottle is designed to be grasped comfortably by an average hand, providing a secure hold and balanced weight distribution when lifting. Engaging with this area ensures stability and effective control during the lifting motion. \\
    Use & To use a bottle, you typically interact with the cap or lid, which is located at the top of the bottle. The cap may have a cylindrical or slightly domed shape and is usually designed to fit snugly over the bottle's neck to create a seal. It may feature ridges or grooves to provide grip for easier twisting. The geometric characteristics include a circular cross-section and a diameter that matches the bottle's opening. The cap's role in completing the task involves being twisted open or closed to either access the contents of the bottle or securely seal them to prevent spillage or contamination. \\
    \bottomrule
    \end{tabular}
    \label{tab:cate_bottle}
\end{table*}

\begin{table*}[b]
    \caption{The generated category-level description of Cylinder\_bottle.}
    \centering
     \renewcommand{\arraystretch}{1.5}
    \begin{tabular}{c p{0.8\textwidth}}
    \toprule
    Task type & \multicolumn{1}{c}{Description} \\
    \midrule
    Hold&	To hold a cylinder\_bottle, the recommended part to interact with is the main cylindrical body of the bottle. The geometric and shape characteristics of this part include a consistent circular cross-section and a comfortable, elongated curvature that fits well within the grasp of a human hand. The cylindrical surface provides an even and stable area for the fingers to wrap around, ensuring a secure grip. This part plays a crucial role in completing the task of holding the bottle by providing structural integrity and an ergonomic design that allows for easy and efficient handling.\\
    Liftup&	To complete the task ``liftup" with the object ``cylinder\_bottle", you should interact with the middle to upper portion of the bottle's body. This part of the cylinder bottle typically features a smooth, rounded, cylindrical shape that provides a stable and ergonomic grip. Its geometric characteristics allow for an even distribution of force when lifting, ensuring better control and reducing the risk of slippage. By holding this part, you can effectively grasp the bottle, maintain balance, and successfully lift it upwards.\\
    Use&	For the task of ``use" involving a ``cylinder\_bottle", the part of the object to interact with is the bottle cap or lid, typically located at the top of the cylindrical body. Geometrically, the cap is usually circular in shape, matching the diameter of the bottle's opening, and is often ribbed or textured to provide grip. The cylindrical bottle itself has a smooth, elongated body, with a cylindrical shape that ensures stability and ease of handling. The role of the cap in completing the task of ``use" is to be twisted or flipped open to access the contents of the bottle, allowing the user to pour, dispense, or consume the liquid inside.\\

    \bottomrule
    \end{tabular}
    \label{tab:cate_cylinder}
\end{table*}

\begin{table*}[b]
    \caption{The generated category-level description of Trigger\_sprayer.}
    \centering
     \renewcommand{\arraystretch}{1.5}
    \begin{tabular}{c p{0.8\textwidth}}
    \toprule
    Task type & \multicolumn{1}{c}{Description} \\
    \midrule
   Hold&	To hold a trigger sprayer, one should interact with the handle or grip, which is typically an ergonomically designed part located at the base of the sprayer near the trigger mechanism. The handle is usually elongated and contoured to fit comfortably in the hand, often with a slightly curved or angled surface to provide a secure and stable grip. The geometric shape of the handle often features indentations or textured surfaces to prevent slipping. Its role in completing the task of holding is to ensure that the user can maintain control and stability of the sprayer during use, facilitating effective and efficient operation.\\
    Liftup&	To perform the task ``liftup" with a ``trigger\_sprayer", interact with the body of the sprayer where it connects to the bottle. This part is typically a cylindrical or slightly tapered shape, designed to fit comfortably in the hand. It provides a stable grip surface, ensuring secure lifting and control of the entire sprayer-bottle assembly. The geometric design aids in evenly distributing the weight of the liquid, making it easier to lift the sprayer efficiently and safely from one place to another.\\
    Use&	To use a trigger sprayer, interact with the trigger, typically located beneath the nozzle and above the handle. The trigger is a small, curved lever designed to fit comfortably against the index and middle fingers. It usually extends outward, allowing for easy squeezing motion. Typically made of plastic, the trigger's shape is ergonomically contoured to facilitate a consistent grip and efficient pumping action. Pulling the trigger activates the internal pump mechanism, which draws liquid up through a straw and expels it through the nozzle as a spray, thus enabling the completion of the spraying task.\\
    \bottomrule
    \end{tabular}
    \label{tab:cate_trigger}
\end{table*}

\begin{table*}[b]
    \caption{The generated part-level description of Bottle.}
    \centering
     \renewcommand{\arraystretch}{1.5}
    \begin{tabular}{c p{0.6\textwidth}}
    \toprule
    part\_lables[object\_category] & \multicolumn{1}{c}{D[part\_label]} \\
    \midrule
    Body&	The body of the bottle is typically the largest and most voluminous part, often cylindrical or slightly curved in shape, serving as the primary container for liquids. It may exhibit a consistent cross-sectional diameter or subtly taper toward the base, aiding stability when placed on flat surfaces. In hand-object interactions, the body provides the main grip area for the user. Its surface characteristics, such as texture and contour, are designed to offer a secure hold, preventing slippage and facilitating easy handling during pouring or drinking.\\
    Lid	&The lid of the bottle is often smaller and has a circular or disc-like geometry that typically matches the bottle's opening. The lid may have threading or snapping mechanisms to secure it onto the neck of the bottle, ensuring a tight seal to prevent spills and maintain the contents' freshness. In hand-object interaction, the lid is the primary component manipulated during opening and closing. It may feature ridges or a textured grip to aid in this process, requiring twisting or pressing motions.\\
    Body Surface&	Body Surface: The body surface encompasses the exterior texture and contouring of the bottle's main section. This surface can range from smooth to ridged and might include ergonomically designed indentations to improve grip. Texturing on the body surface enhances friction between the hand and the bottle, which is crucial when the object is wet or slippery. In hand-object interaction, the body surface's tactile properties affect how securely the bottle can be held during transport or consumption.\\
    Neck&	The neck of the bottle is a narrower section that connects the body to the rim and lid. Its shape is usually cylindrical and may taper slightly towards the top, facilitating a gradual transition between the wider body and narrower opening. The neck serves both a functional and aesthetic purpose, helping to guide liquid flow while pouring and affecting how the bottle balances in the user's hand. When interacting with the bottle, the neck is often used as a secondary gripping point, especially when pouring.\\
    Rim&	The rim refers to the thin edge or border at the very top of the bottle's opening. Geometrically, it is often slightly raised or rounded to create a smooth boundary at the top of the neck. The rim ensures a clean pour by guiding the liquid flow and can also provide a sealing surface for the lid. In terms of hand-object interaction, the rim plays a subtle but important role in drinking directly from the bottle and ensuring a pleasant mouthfeel. When capping the bottle, it contributes to a secure closure by interacting precisely with the lid's corresponding surface.\\

    \bottomrule
    \end{tabular}
    \label{tab:part_bottle}
\end{table*}

\begin{table*}[b]
    \caption{The generated part-level description of Cylinder\_bottle.}
    \centering
     \renewcommand{\arraystretch}{1.5}
    \begin{tabular}{c p{0.6\textwidth}}
    \toprule
    part\_lables[object\_category] & \multicolumn{1}{c}{D[part\_label]} \\
    \midrule
    Body&	The body of a cylinder\_bottle is the largest and most prominent part. Geometrically, it is a cylindrical shape, characterized by a circular cross-section and a uniform diameter along its height. The body’s axis is typically aligned vertically, and it provides the primary volume for containing liquids or other contents. In hand-object interaction, the body serves as the main gripping area. Its cylindrical shape allows for a firm and ergonomic grip with one or both hands, depending on the size of the bottle.\\
    Cap&	The cap of the cylinder\_bottle is a detachable part that seals the bottle, located at the very top of the head or neck. It can come in various shapes, such as screw caps, snap-on lids, or corks, each designed to secure the contents and prevent spillage. The cap may be cylindrical, slightly domed, or have ergonomic features such as ridges for better grip. In hand-object interaction, the cap is crucial for accessing the bottle’s contents. It requires manual dexterity to open and close, with its design ensuring ease of use and secure sealing to maintain the integrity of the contents.\\
    Head&	The head of the cylinder\_bottle is the uppermost part, leading to the neck. It generally includes a narrowing transition from the body to the neck, often featuring a slight tapering or rounded design. This part ensures a secure fit for the cap and helps to guide the flow of the liquid during pouring. In terms of hand-object interaction, the head is not usually the primary area for gripping. However, it plays a crucial role when the bottle is tilted for pouring, requiring stability and support from the user's secondary hand.\\
    Neck&	The neck of the cylinder\_bottle is the part situated between the head and the cap. Geometrically, it is narrower than the body and head, often cylindrical but with a smaller diameter. The neck acts as a conduit for controlled flow during pouring or drinking and plays a structural role in accommodating the cap. In hand-object interaction, the neck is often used as a secondary gripping point, especially when precision pouring is required. Its design allows for a comfortable hold that aids in the control of liquid flow.\\
    Surface&	The surface refers to the exterior area encompassing the external parts of the cylinder bottle, which include the body, head, neck, and cap. Geometrically, the surface is continuous and smooth, though it may feature textural elements (such as ridges or patterns) to improve grip. In hand-object interaction, the surface characteristics are crucial as they provide tactile feedback and friction, which help prevent slipping. Good surface design enhances the user's confidence during activities like opening the cap or holding the bottle securely, especially when the surface is wet or oily.\\

    \bottomrule
    \end{tabular}
    \label{tab:part_cylinder}
\end{table*}

\begin{table*}[b]
    \caption{The generated part-level description of Trigger\_sprayer.}
    \centering
     \renewcommand{\arraystretch}{1.5}
    \begin{tabular}{c p{0.6\textwidth}}
    \toprule
    part\_lables[object\_category] & \multicolumn{1}{c}{D[part\_label]} \\
    \midrule
Body&	The body of the trigger sprayer is the central structure connecting the head to the nozzle and other components. It serves as a housing for the internal pumping system, which can include pistons, seals, and springs. Geometrically, the body is often cylindrical or square-like with structural reinforcements to withstand the internal pressures during operation. The body also features attachment points that secure it to the container. In hand-object interaction, the body provides the main structural support and is often where the user’s grip is stabilized, allowing controlled and directed force application when the trigger is pressed.\\
Container surface&	The container surface refers to the external area of the bottle or container attached to the trigger sprayer. This part has a generally cylindrical form but can also be rectangular or sculpted for ergonomic handling. The surface might include textured grips, volume measurements, or branding information. For hand-object interaction, the container surface provides the main gripping area, allowing the user to secure the sprayer firmly while applying force on the trigger and directing the spray.\\
Head&	The head of the trigger sprayer is typically a small, ergonomically shaped component designed to fit the human hand comfortably. It features grooves or textures to enhance grip, ensuring that the user can easily exert pressure when using the sprayer. The head houses the internal mechanisms that control the flow of liquid, often including a spring and valve assembly. During hand-object interaction, the head is the primary point of contact where the user applies force to activate the spray mechanism, thus playing a crucial role in the sprayer's functionality.\\
Neck&	The neck of the container is the transitional section between the body and the container. It is usually narrower than the body and features threads or latches that allow the sprayer body to be securely attached. Geometrically, the neck often includes structural reinforcements to handle the stress of attaching and detaching the sprayer head. In hand-object interaction, the neck is pivotal for ensuring a secure connection between the sprayer mechanism and the liquid container. Users engage with the neck primarily when installing or removing the sprayer from the container.\\
Nozzle&	The nozzle, usually located at the front end of the head, is a small, adjustable component often conically shaped to help focus the liquid spray into a stream or mist. Its geometric configuration allows for rotation to adjust spray settings, requiring fine-tuned manipulation by the fingers. This part is crucial for directing the outflow of the liquid, dictating the pattern, range, and concentration of the spray, thus aiding users in targeting specific areas with precision.\\
    \bottomrule
    \end{tabular}
    \label{tab:part_lotion}
\end{table*}
\clearpage
\end{document}